\documentclass[sn-mathphys-num]{sn-jnl}

\usepackage{amsmath,amssymb,amsfonts}%
\usepackage{amsthm}%
\usepackage{mathrsfs}%
\usepackage[title]{appendix}%
\usepackage{xcolor}%
\usepackage{textcomp}%
\usepackage{manyfoot}%
\usepackage{booktabs}%
\usepackage{algorithm}%
\usepackage{algorithmicx}%
\usepackage{algpseudocode}%
\usepackage{listings}%
\usepackage{tabularx}
\usepackage{adjustbox}

\usepackage[justification=centering]{caption} 
\usepackage{amsfonts}
\usepackage{subfigure}
\usepackage{multirow}
\usepackage{booktabs}
\usepackage{epstopdf}
\AppendGraphicsExtensions{.tif}
\usepackage{graphicx}%

\newlength{\patchfivewidth} 
\newlength{\patchsixwidth} 
\newlength{\patchsevenwidth} 
\begin{document}
\bibliographystyle{plain}
\title[]{Support-Query Prototype Fusion Network for Few-shot Medical Image Segmentation}


\author[1,3]{\fnm{Xiaoxiao} \sur{Wu}}\email{wxx\_52v@126.com}

\author*[1,3]{\fnm{Zhenguo} \sur{Gao}}\email{gaohit@sina.com}

\author[1,3]{\fnm{Xiaowei} \sur{Chen}}\email{932197321@qq.com}
\equalcont{These authors contributed equally to this work.}

\author[1,3]{\fnm{Yakai} \sur{Wang}}\email{1045923522@qq.com}
\equalcont{These authors contributed equally to this work.}

\author[1,3]{\fnm{Shulei} \sur{Qu}}\email{13562506757@163.com}
\equalcont{These authors contributed equally to this work.}

\author[2,3]{\fnm{Na} \sur{Li}}\email{1831101596@qq.com}
\equalcont{These authors contributed equally to this work.}

\affil*[1]{\orgdiv{Department of Computer Science and Technology}, \orgname{Huaqiao University}, \orgaddress{\street{No.668 Jimei Avenue}, \city{Xiamen}, \postcode{361021}, \state{Fujian}, \country{China}}}
\affil[2]{\orgdiv{Department of Mechanical Engineering and Automation}, \orgname{Huaqiao University}, \orgaddress{\street{No.668 Jimei Avenue}, \city{Xiamen}, \postcode{361021}, \state{Fujian}, \country{China}}}
\affil[3]{\orgdiv{}\orgname{Key Laboratory of Computer Vision and Machine Learning of Fujian Provincial Universities}, \orgaddress{\street{}\city{Xiamen}, \postcode{361021}, \state{Fujian}, \country{China}}}


\abstract{In recent years, deep learning based on Convolutional Neural Networks (CNNs) has achieved remarkable success in many applications. However, their heavy reliance on extensive labeled data and limited generalization ability to unseen classes pose challenges to their suitability for medical image processing tasks. Few-shot learning, which utilizes a small amount of labeled data to generalize to unseen classes, has emerged as a critical research area, attracting substantial attention. Currently, most studies employ a prototype-based approach, in which prototypical networks are used to construct prototypes from the support set, guiding the processing of the query set to obtain the final results. While effective, this approach heavily relies on the support set while neglecting the query set, resulting in notable disparities within the model classes. To mitigate this drawback, we propose a novel Support-Query Prototype Fusion Network (SQPFNet). SQPFNet initially generates several support prototypes for the foreground areas of the support images, thus producing a coarse segmentation mask. Subsequently, a query prototype is constructed based on the coarse segmentation mask, additionally exploiting pattern information in the query set. Thus, SQPFNet constructs high-quality support-query fused prototypes, upon which the query image is segmented to obtain the final refined query mask. Evaluation results on two public datasets, SABS and CMR, show that SQPFNet achieves state-of-the-art performance.
}


\maketitle

\section{Introduction}\label{sec1}
Semantic segmentation is a cornerstone task in computer vision, aimed at pixel-level classification of images by assigning each pixel to a specific class, such that the image is segmented into several regions. Convolutional Neural Networks (CNNs) have achieved remarkable success in natural image analysis, as well as a widespread application in the medical domain. Semantic segmentation of medical images plays a pivotal role in advancing medical image analysis, enabling the automatic identification and characterization of anatomical structures and pathological regions, including tumors, organs, blood vessels, and lesions. This in turn facilitates accurate diagnosis, treatment planning, and disease monitoring. 

However, semantic segmentation of medical images presents unique challenges compared to natural image segmentation, as medical images often exhibit low contrast, high variability, heterogeneity, and noise, thereby intensifying the complexity of segmentation tasks. Furthermore, the success of CNNs rely on large amount of well-annotated training data, otherwise their performance will be seriously degraded. However, being of more paramount importance for medical image analysis, precise annotations are seriously limited as requiring expensive labor efforts of high specialized technicians. Hence, the scarcity of annotated medical image datasets also poses a significant challenge to medical image analysis. 

To overcome the dataset scarcity challenge, few-shot learning is increasingly attracting research efforts, which aims to quickly learn the underlying patterns from limited annotated images (in support set) and well adapt to new images (in query set). Few-Shot Segmentation (FSS) is a typical paradigm of few-shot learning. FSS consider images from some classes. An image with its corresponding ground truth segmentation mask is called an annotated image, and the mask is called the annotation of the image. A pair of an image and its mask is called an image-annotation pair. As an annotated image is always accompanied with its mask, the phrase \textit{annotated image} is usually referred to the image-annotation pair by default for description clarity. The actual meaning can be distinguished from the context.

Most existing FSS models in this domain rely on prototypical networks, which extract class prototypes from the support set for each class and then can be well applied to segment the query set. However, this approach often results in an over-reliance on the support set, potentially overlooking valuable information in the query set. Recent research have explored various methods to address this issue. For example, Wang \textit{et al.}\cite{b1} focused on adapting to the differences between the support and query images by including alignment loss. Ding \textit{et al.}\cite{b2} leveraged pixel-level relationships between the query and support images using periodic similarity attention module. Additionally, method proposed by Sun \textit{et al.}\cite{b3} exploited connections between the support and query sets, while that of Shen \textit{et al.}\cite{b4} allowed continuous prototype adaptation even in the testing stage, aiming to refine segmentation predictions. Despite of these efforts, existing approaches often under-exploit the information in the query set during training. To fill this gap, we incorporate query features into the prototype construction, enabling to obtain a support-query fused prototype by utilizing information from both the support set and the query set. Our approach involves segmenting the foreground area of the support image into parts and constructing support prototypes for them. These support prototypes serve as the basis for building a comprehensive class prototype, not only for the support set but also for the query set. Finally, we aggregate the heterogeneous prototypes to obtain the final support-query fused prototype, which is used to obtain a refined segmentation result of the query image. We call our model as Support-Query Prototype Fusion Network (SQPFNet). Extensive simulation results on public datasets confirms that, by leveraging both support and query set information effectively, our SQPFNet approximately obtain the state-of-the-art performance. 

\begin{figure*}[t]
\setlength{\belowcaptionskip}{5mm}
\centerline{\includegraphics[width=\textwidth]{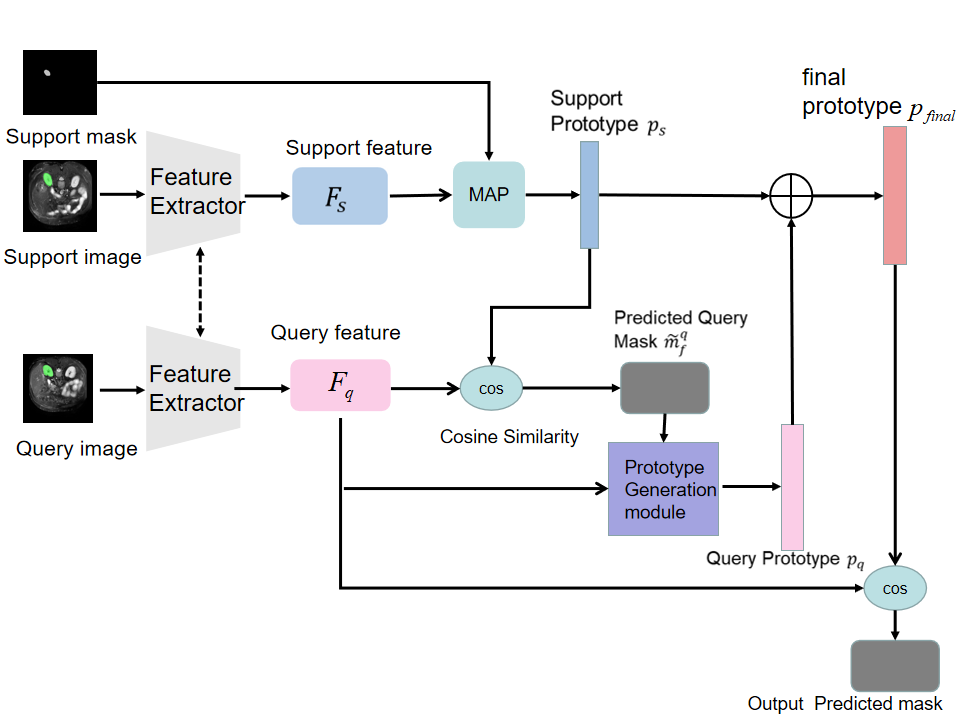}}
\captionsetup{justification=raggedright}
\caption{Network structure of SQPFNet.}
\label{fig1}
\end{figure*}

The contributions of this paper are summarized as follows:

\begin{itemize}

\item{We proposed an new support-query prototype fusion approach to jointly learns information from both the support images and the query image. It achieves this goal by constructing a refined query prototype via exploiting the query image to refine the comprehensive prototype which combines the heterogeneous prototypes generated from the segmentation parts in foregrounds of the support images.}

\item{We proposed SQPFNet following our support-query prototype fusion approach for few-shot segmentation of medical images.}

\item{Extensive experiments were carried out on two different medical datasets, and the simulation results verified the superiority of our SQPFNet.}

\end{itemize}

\section{Related Work}\label{sec2}
\subsection{Medical Image Segmentation}
UNet proposed by Ronneberger \textit{et al.}\cite{b5} presents a symmetrical encoder-decoder architecture with skip connections. In the encoder, features are progressively extracted using convolutional operations, while the decoder combines low-level encoder features with high-level ones via skip connections, facilitating the fusion of features across different levels. UNet is regarded as a significant milestone in medical image segmentation, serving as a foundational framework for subsequent research endeavors. 

To leverage 3D medical image data more effectively, Cicek \textit{et al.}\cite{b6} replaced UNet's convolutions with 3D convolutions. Milletari \textit{et al.}\cite{b7} brought residual connections into UNet to address the gradient vanishing issue, while VoxResNet proposed in \textit{et al.}\cite{b8} augments UNet with residual structures, enhancing its capacity to harness deep features. Additionally, Xiao \textit{et al.}\cite{b9} proposed weighted-ResUNet, which incorporates weighted attention mechanisms and skip connections to deal with thin blood vessel segmentation and optic disc recognition in retinal vascular segmentation tasks. Although these 3D networks achieve improved segmentation performance, their increased model sizes lead to high computation costs. 

For complex segmentation tasks, model cascading offers a solution. In this approach, multiple segmentation models are integrated using a cascaded structure, where each model is dedicated to solving a specific sub-task and generating outputs to feed subsequent models. For instance, Christ \textit{et al.}\cite{b10} proposed a cascade FCN\cite{b11} approach for liver and lesion segmentation, while Kaluva \textit{et al.}\cite{b12} devised DenseNet, which is a two-stage cascade segmentation method for liver and tumor segmentation. Moreover, Yan \textit{et al.}\cite{b13} introduced a multi-scale encoder-decoder cascade model to leverage automatic context integration of multi-scale information, in order to meet the demands of higher-resolution segmentation tasks. Although these networks have achieved promising results, they often require substantial amounts of annotated data, posing challenges in clinical medical image analysis.

\subsection{Few-shot Learning}
Deep learning has achieved remarkable success across various domains, including image recognition, natural language processing, and video analysis. However, the effectiveness of deep learning models heavily depends on large labeled datasets, which are often scarce in practical applications. Consequently, research efforts have shifted towards few-shot learning, aiming to achieve satisfactory results with limited labeled data. Currently, few-shot learning has three approaches: model-based learning, optimization-based learning, and metric-based learning. 

In model-based approach, techniques like Neural Turing Machines (NTM)\cite{b14} are employed to incorporate short-term memory via external storage and long-term memory through gradual weight updates. The Meta Network\cite{b15} utilizes a fast-weight mechanism generated from training gradients to adaptively update parameters during inference. 

In optimization-based approach, Ravi \textit{et al.}\cite{b16} recognized that gradient-based optimization algorithms, especially of non-convex problems with numerous hyper-parameters, leads to slow convergence. To deal with this issue, they proposed a model which learns to initialize and update the parameters of classifier networks on new tasks by learning parameter updating rules. 

Metric-based approach addresses over-fitting in low-data scenarios by employing non-parametric methods, such as nearest neighbors or K-Means, to construct end-to-end classifiers. Siamese networks\cite{b17} leverages cosine similarity to assess feature similarity, where similarity is computed during training either between an anchor point and the positive/negative samples or between pairs of vectors. The prototypical network proposed in \cite{b18} projects samples into a shared space and compute class prototypes by averaging samples within each class, where distances between samples and class prototypes are computed during testing using Euclidean distance, and probabilities are calculated using the softmax function. The class with the highest score is then regarded as the sample's category. 

\subsection{Few-shot Semantic Segmentation}

In recent years, significant progress has been made in the field of few-shot semantic segmentation. Shaban \textit{et al.}\cite{b19} pioneered this area by introducing the concept of few-shot segmentation. Their approach involves a two-arm structure comprising a conditioning arm and a segmentation arm. The conditioning arm processes annotated support images to generate feature representations, whereas the segmentation arm executes the segmentation task. To further leverage support set information, Wang \textit{et al.}\cite{b1} introduced an alignment strategy, where an alignment loss is employed to swap the positions of a support image and a query image during training, enabling the use of the query image to predict the support image. This approach helps narrow the gap between the support set and the query set. Additionally, Ouyang \textit{et al.}\cite{b20} and Hansen \textit{et al.}\cite{b21} respectively introduced the concepts of superpixel and supervoxel for facilitating segmentation tasks. Shen \textit{et al.}\cite{b4} proposed a prototype refinement module aimed at continuously updating acquired class prototypes during testing, leading to more precise predictions. Feng \textit{et al.}\cite{b22} introduced a spatial branch to incorporate spatial information into segmentation tasks, integrating this with prototype similarity for improved predictions.

Most existing methods focus on facilitating interaction between support sets and query sets. Ding \textit{et al.}\cite{b2} enhanced the non-local block by selectively boosting query and support features through extensive similarity comparison between query and support pixels. Sun \textit{et al.} \cite{b3} proposed a global correlation module to encourage feature clustering, enhancing similarity between support and query images. Wei \textit{et al.}\cite{b24} introduced a network for embedding anatomical knowledge, where anatomical prior knowledge from support images is initially embedded into the model, and then a similarity-guided strategy is utilized to align query and support sets. 

\section{SQPFNet for Few-Shot Segmentation}
\label{sec3}
\subsection{Problem Definition}
Given a dataset of annotated images from a set of $n_{train}$ classes in $C_{train}{=}\{c_1,c_2,\ldots,c_{n_{train}}\}$ classes, FSS aims to quick adaption to new classes $C_{test}{=}\{c_1,c_2,\ldots,c_{n_{test}}\}$, such that it can well segment images from classes $C_{test}$ when exposed to only a few annotated sample images from $C_{test}$. Here $C_{train}{\cap}C_{test}{=}\emptyset$ is required. An FSS model is trained on the training set and evaluated on the test set. 

In FSS, the training and testing are performed in an episodic manner ~\cite{ref_Vinyals2016}. In each episode of the training stage, a set of $N$ classes are sampled from $C_{train}$, and $K$ annotated images from each of the $N$ classes are sampled from the dataset to create a support set $\mathcal{S}$. Let $C_S$ be the set of $N$ selected classes in this episode. Meanwhile, a query set $\mathcal{Q}$ is also created by selecting an annotated image from a random class $c{\in}C_S$. The testing stage is identical to the training stage, except that the $N$ classes are selected from $C_{test}$. We refer an image in the support set $\mathcal{S}$ as a support image, and that in the query set $\mathcal{Q}$ as a query image. This setting is named as $N$-way-$K$-shot setting. To be specific, in an $N$-way-$K$-shot setting, we can denote the set $\mathcal{S}$ as $\mathcal{S}{=}\{(x^s_{i,k},m^s_{i,k}|i{\in}\{1,2,\ldots,N\}, k{\in}\{1,2,\ldots,K\}\}$ where $x^s_{i,k}$ represents the $k$-th image in the support set $\mathcal{S}$ from the $i$-th class in $c{\in}C_S$, and $m^s_{i,k}$ represents the corresponding ground truth segmentation mask. Similarly, we can denote the query set $\mathcal{Q}$ as $\mathcal{Q}=\{(x^q_j,m^q_j)\}$, where $x^q_j$ and $m^q_j$, $j{\in}C_{S}$, represent the query image and its corresponding ground truth mask, respectively. The model's task is to learn from the support set $\mathcal{S}$ and generalize its knowledge to accurately segment the image in the query set $\mathcal{Q}$.

\subsection{SQPFNet Structure Overview}
Our SQPFNet adopts the prototypical network model approach. For facilitating the readers, the structure of SQPFNet is re-illustrated in Figure~\ref{fig1}.

\begin{figure*}[!htpb]
\setlength{\belowcaptionskip}{5mm}
\centerline{\includegraphics[width=\textwidth]{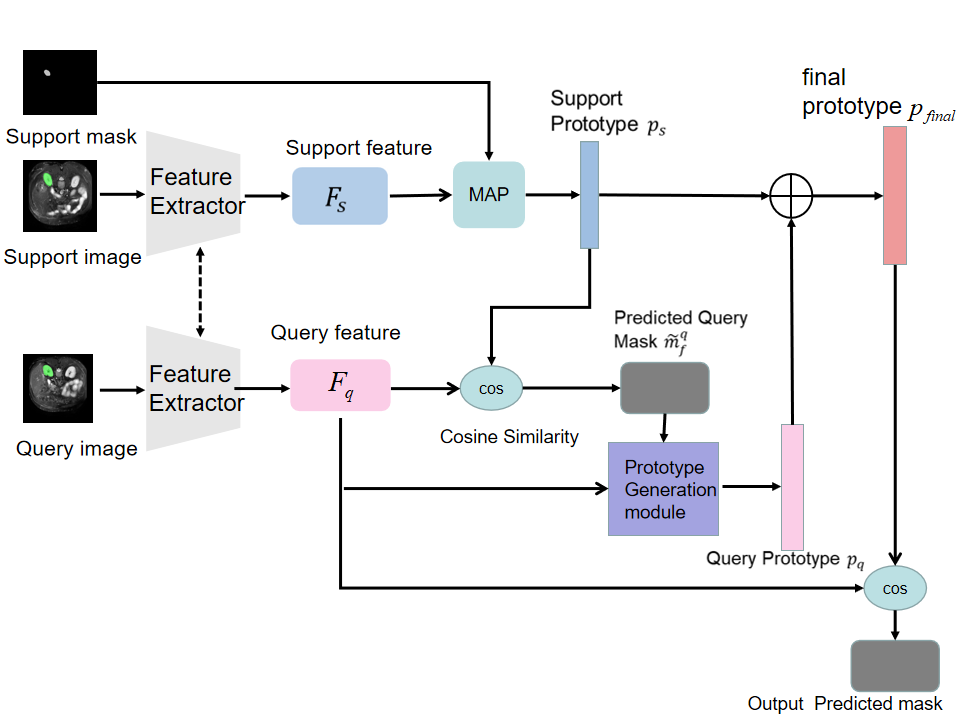}}
\captionsetup{justification=raggedright}
\caption{Network structure of SQPFNet.}
\label{fig1}
\end{figure*}

The process and the key idea of SQPFNet is as follow. Initially, it extracts features from both the support and query images using a feature extractor. Then, it constructs a class prototype for the support set (called as support prototype) via the prototype generation module. A class prototype for the query set (called as query prototype) is then generated by exploiting guiding hues from a coarse query mask predicted using the support prototype. The support prototype and query prototype are fused to yield the final query prototype. Ultimately, the query image is segmented by leveraging the final query prototype, leading to a refined query mask as the segmentation result of the query image.

\subsection{Support-Query Prototype Fusion}
Many existing medical few-shot segmentation models rely on prototypical networks. The essence of a prototypical network lies in acquiring the class information from the support sets and representing them as prototypes. When both the query sets and prototypes are projected into a vector space, an image in the query set is categorized based on its proximity to the prototypes. This metric-based learning approach is both simple and effective, making it particularly suitable for medical image segmentation due to the significant disparity between the foreground and the background pixels. 

Firstly, the model utilizes a feature extractor ${f_\theta }$ to extract support features ${F_s}$ and query features ${F_q}$. The dimensions of the feature map are denoted as ${H'}$ for height, ${W'}$ for width, and ${C}$ for the number of channels, respectively. Following the state-of-the-art approach, we employ the ResNet101 model pre-trained on the COCO dataset as the feature extractor. The extracted feature is then reduced to 1/4 of the input size. In practical implementation, we utilized the torchvision library's \textit {deeplabv3\_resnet101} model, yet remove the ASPP (Atrous Spatial Pyramid Pooling) module.

We employ a prototype-based approach for few-shot segmentation tasks within this domain. Due to the divergence between the images in the support set and the query set, constructing a single prototype for the entire support set may not adequately capture the class information of the support images. Hence, we opt to partition a support image into different regions and establish a prototype for each region.

Initially, we identify the foreground region ${R_n}$ based on the support image ${x_s}$ and its corresponding mask ${m_s}$. We then partition both the foreground region ${R_n}$ and its mask ${M_n}$ into ${N_r}$ segments, denoted as $\mathcal{R}_n{=}\{r_i\}_{i{=}1}^{{N_r}}$ and $\mathcal{M}_n{=}\{u_i\}_{i{=}1}^{N_r}$, respectively. For each segment, we derive a regional prototype by conducting pixel-level Mask Average Pooling~(MAP) on the support features (features in support images) and masks within that specific segment. The amalgamation of all regional prototypes yields the ultimate support prototype ${p_s}$, which is calculated using Eq.~\eqref{eq_ps} where ${\odot}$ denotes Hadamard product.

\begin{equation}
\label{eq_ps}
{p_s} =\frac{1}{N_r}\sum_{i{=}1}^{N_r}\text{MAP}({F_s},{u_i}) = 
\frac{1}{N_r}\sum_{i{=}1}^{N_r}{\cdot}
\frac{\sum\nolimits_{x,y} {\left(F_s(x,y){\odot}u_i(x,y)\right)}}{\sum\nolimits_{x,y}u_i(x,y)}
\end{equation}

When calculating the average mask, we sample the support mask of the region ${u_i}$ to the size of the support feature ${F_s}$, which is ${(C, H', W')}$.

After obtaining the support prototype ${p_s}$, the initial query mask ${\tilde m^q}$ is derived by computing the cosine similarity between the support prototype ${p_s}$ and the query features ${F_q}$ as Eq.~\eqref{eq_mq}, where $\cos$ denotes cosine similarity and ${F_q}$ denotes query features.

\begin{equation}
\label{eq_mq}
 {\tilde m^q} = \text{softmax} (\cos ({F_q},{p_s})).
\end{equation}

For few-shot segmentation tasks, a common challenge arises from intra-class gaps. Even though the support set and the query set belong to the same category, variations in imaging sensors can lead to differences, even among organs of the same category. Direct utilization of the support set's extracted prototypes may bring excessive impacts from the support set, resulting in inaccurate segmentation of query image. To address this issue, we propose building query prototypes akin to how supporting prototypes are constructed. The key difference lies in the absence of a genuine query mask for the query prototype. To compensate for it, we suggest utilize the query mask ${\tilde m^q}$ derived from the support prototype and selecting a portion of it as the ground-truth query mask. Despite potential biases in the query mask obtained from the support prototype, it can be viewed as a coarse query mask, containing partially relevant information. The query prediction generated based on the support prototype is ${\tilde m^q}$, it is partitioned into two components: the foreground prediction mask ${\tilde m_f^q}$ and the background prediction mask ${\tilde m_b^q}$, where
\begin{equation}
    \tilde{m}_b^q = 1 - \tilde{m}_f^q.
\end{equation}

As previously discussed, we employ a partial filtering mechanism for the query prediction mask utilizing support prototypes. In essence, we adopt a hard selection method to establish thresholds for both the foreground prediction mask and the background prediction mask. Typically, the threshold for the foreground ${T_f}$ is set relatively high, whereas the threshold for the background ${T_b}$ is set relatively low. Empirically, we set the background threshold  to 0.5, while the formula for computing the foreground threshold ${T_f}$ is as follows
\begin{equation}
{T_f} = \sum\nolimits_{x,y} {\frac{\max\left(\tilde m_f^q(x,y)\right) + \text{mean}\left(\tilde m_f^q(x,y)\right)}{2}}. 
\end{equation}

After foreground threshold ${T_f}$ and background threshold are obtained, the query prediction mask after hard selection ${m'}$ can be expressed as Eq.~\eqref{eq_mfp} and Eq.~\eqref{eq_mbp}, where the subscripts ${f}$ and ${b}$ are used to indicate foreground and background, respectively.
\begin{equation}
\label{eq_mfp}
m_f' = \left\{ 
\begin{array}{lr} 
1, &\hspace{5mm} \text{if } \tilde{m}_f^q{\ge}{T_f};\\
0, &\hspace{5mm}\text{otherwise}.
\end{array} \right. 
\end{equation}

\begin{equation}
\label{eq_mbp}
m_b' = \left\{ 
\begin{array}{lr}
1,  &\hspace{5mm}\text{If }\tilde{m}_b^q{\geq}T_b;\\
0,  &\hspace{5mm}\text{otherwise}.
\end{array}
\right.
\end{equation}

We consider the filtered query prediction mask ${m'}$ as the ground-truth and proceed to obtain the query prototype ${p_q}$ in a similar manner to acquiring the support prototype. Specifically, we perform average pooling on the combination of the query feature and the query mask after hard selection ${m'}$ using Eq.~\eqref{eq_pq}.

\begin{equation}
\label{eq_pq}
{p_q} = \frac{{\sum\nolimits_{x,y} {{F_q}(x,y) \odot m'(x,y)} }}{\sum\nolimits_{x,y}{m'(x,y)}}.
\end{equation}

We fuse the support prototype ${p_s}$ with the query prototype ${p_q}$ to obtain the final prototype ${p_\text{final}}$ using Eq.~\eqref{eq_prototype_fuse}, where $\alpha$ and $\beta$ are two parameters for controlling the relative importantce of the two prototypes. 

\begin{equation}
\label{eq_prototype_fuse}
{p_\text{final}}{=}\alpha{\cdot}{p_s}{+}\beta{\cdot}{p_q}.
\end{equation}

Utilizing the final prototype ${p_\text{final}}$, we obtain the query foreground prediction mask ${\hat m_f^q}$ using Eq.~\eqref{eq_mfq}.
\begin{equation}
\label{eq_mfq}
\hat m_f^q = \text{softmax}(\cos(p_{final},F_q))
\end{equation}

\section{Experiment}
\subsection{Dataset}
To validate the effectiveness of our method, we conducted experiments on two representative publicly available datasets: SABS \cite{b25} and CMR\cite{b26}. The SABS dataset originates from the Multi-Atlas Abdominal Label Challenge released by MICCAI2015. It comprises CT scans of the abdomen with multi-organ segmentation, featuring organs such as the pancreas, kidneys (left and right), gallbladder, adrenal glands (left and right), liver, and inferior vena cava, among others. This dataset consists of 30 3D CT scans. The CMR dataset, on the other hand, is a cardiac segmentation dataset published by MICCAI in 2019. It encompasses 35 scans, each with approximately 13 slices per scan. 

To ensure fair comparison with other methods, we employ a consistent setting by adjusting all images to either 2D axial or 2D short-axis. Specifically, for the SABS dataset, we convert it to a 2D axial representation, while for the CMR dataset, we transform it to a 2D short-axial format. Subsequently, all adjusted images are resize to 256×256 size. Focusing on the SABS dataset, akin to prior approaches, we concentrate our analysis on the model's performance concerning the kidneys, liver, and spleen. Furthermore, all experiments follow a five-fold cross-validation methodology.

\begin{table}[t]
\centering
\captionsetup{justification=raggedright}
\caption{Comparison results of SQPFNet with other methods on the SABS dataset.}
\begin{tabularx}{0.95\textwidth}{@{}c|c|cccc|c}
\toprule
\textbf{Settings} & \textbf{Methods} & \textbf{LK}    & \textbf{RK}    & \textbf{Spleen} & \textbf{Liver} & \textbf{Mean Dice Score$\uparrow$}  \\ 
\midrule
\multirow{6}{*}{1}   & ALPNet\cite{b20} & 72.36   & 71.81  & 70.96  & 78.29  & 73.35 \\
    & ADNet\cite{b21}  & 68.76   & 61.18  & 68.10  & 79.49  & 69.38\\
    & QNet\cite{b4}    & 74.14   & 74.04  & \textbf{\color{red}{78.32}}  &  \color{blue}{80.70}  & \color{blue}{76.80} \\
    & AAS-DCL\cite{b27}& 74.58   & 73.19  & 72.30  & 78.04  & 74.52 \\
    & CRAPNet\cite{b2} & \color{blue}{74.69}   & \color{blue}{74.18}  & 70.37  & 75.41  & 73.66 \\
    & SQPFNet(Ours)     & \textbf{\color{red}{74.88}} & \textbf{\color{red}{75.10}} &  \color{blue}{73.20} & \textbf{\color{red}{84.79}}   & \textbf{\color{red}{77.00}}\\
\midrule
\multirow{6}{*}{2}   & ALPNet\cite{b20} & 63.34  & 54.82  & 60.25  & 73.65  & 63.02\\
    & ADNet\cite{b21}  & 36.52  & 37.60  & 46.10  & 76.25  & 49.12 \\
    & QNet\cite{b4}    & \color{blue}{67.73}  & 62.11  & \color{blue}{62.35}  & \color{blue}{81.45}  & \color{blue}{68.41} \\
    & AAS-DCL\cite{b27}& 64.71  & \textbf{\color{red}{69.95}}  & \textbf{\color{red}{66.36}}  & 71.61 & 68.16\\
    & CRAPNet\cite{b2} & 53.31  & 63.97  & 60.25  & 78.22  & 63.94\\
    & SQPFNet(Ours)          & \textbf{\color{red}{68.93}} & \color{blue}{66.99}   & 62.08  & \textbf{\color{red}{81.47}} & \textbf{\color{red}{69.87}} \\ 
\bottomrule
\end{tabularx}
\label{table1}
\end{table}

\begin{table}[t]
\centering
\captionsetup{justification=raggedright}
\caption{Comparison results of SQPFNet with other methods on the CMR dataset.}
\begin{tabularx}{0.95\textwidth}{c|c|ccc|c}
\toprule
\textbf{Settings} & \textbf{Methods} & \textbf{LV-BP} & \textbf{LV-MYO} & \textbf{RV}    & \textbf{Mean Dice Score$\uparrow$}\\ 
\midrule
\multirow{4}{*}{1}         
& ALPNet\cite{b20}  & 83.99  & \textbf{\color{red}{66.74}}  & \textbf{\color{red}{79.96}} & \color{blue}{76.90} \\
& ADNet\cite{b21}   & 86.62  & 62.41           & 74.93          & 74.65 \\
& Q-Net\cite{b4}    & \color{blue}{89.23}  & 64.42           & 75.70          & 76.45 \\
& SQPFNet(Ours)            & \textbf{\color{red}{89.97}} & \color{blue}{64.96}   & \color{blue}{77.29}  & \textbf{\color{red}{77.41}} \\ \bottomrule
\end{tabularx}
\label{table2}
\end{table}

\subsection{Evaluate Metric}
In accordance with common medical segmentation practices, we employ the average Dice coefficient as an evaluation metric. The Dice coefficient, ranging between 0 and 100, indicates the degree of overlap between the segmented region and the ground truth. A score of 0 implies complete disjointedness between the segmentation result and the actual mask, while a score of 100 signifies perfect alignment between the segmentation result and the ground truth mask.
The Dice score is calculated using the following formula:
\begin{equation}
    {\rm{D}}(A,B) = 2\frac{{|A \cap B|}}{{|A| + |B|}} \cdot 100\% 
\end{equation}

\subsection{Comparison of SQPFNet with State-of-the-Art Methods}

\begin{figure*}[!t]
\setlength{\belowcaptionskip}{2mm}
\centering
\subfigure{
\begin{minipage}[t]{\patchsixwidth}
\centering
{\scriptsize{ALPNet\cite{b20}}}
\end{minipage}
\begin{minipage}[t]{\patchsixwidth}
\centering
{\scriptsize{ADNet\cite{b21}}}
\end{minipage}
\begin{minipage}[t]{\patchsixwidth}
\centering
{\scriptsize{Q-Net\cite{b4}}}
\end{minipage}
\begin{minipage}[t]{\patchsixwidth}
\centering
{\scriptsize{CRAPNet\cite{b2}}}
\end{minipage}
\begin{minipage}[t]{\patchsixwidth}
\centering
{\scriptsize{SQPFNet(Ours)}}
\end{minipage}
\begin{minipage}[t]{\patchsixwidth}
\centering
{\scriptsize{GT}}
\end{minipage}
}
\hspace{-3mm}
\subfigure{
\begin{minipage}[t]{\patchsixwidth}
\centering
\includegraphics[width=\textwidth]{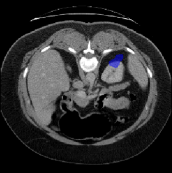}
\end{minipage}%
}%
\subfigure{
\begin{minipage}[t]{\patchsixwidth}
\centering
\includegraphics[width=\textwidth]{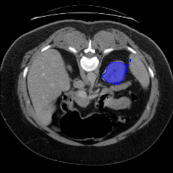}
\end{minipage}%
}%
\subfigure{
\begin{minipage}[t]{\patchsixwidth}
\centering
\includegraphics[width=\textwidth]{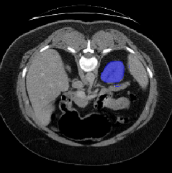}
\end{minipage}%
}%
\subfigure{
\begin{minipage}[t]{\patchsixwidth}
\centering
\includegraphics[width=\textwidth]{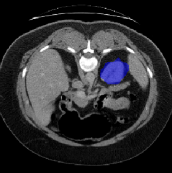}
\end{minipage}%
}%
\subfigure{
\begin{minipage}[t]{\patchsixwidth}
\centering
\includegraphics[width=\textwidth]{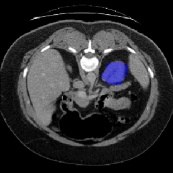}
\end{minipage}%
}%
\subfigure{
\begin{minipage}[t]{\patchsixwidth}
\centering
\includegraphics[width=\textwidth]{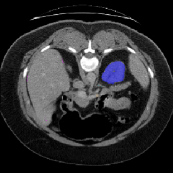}
\end{minipage}%
}%

\subfigure{
\begin{minipage}[t]{\patchsixwidth}
\centering
\includegraphics[width=\textwidth]{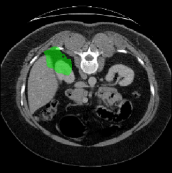}
\end{minipage}%
}%
\subfigure{
\begin{minipage}[t]{\patchsixwidth}
\centering
\includegraphics[width=\textwidth]{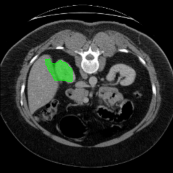}
\end{minipage}%
}%
\subfigure{
\begin{minipage}[t]{\patchsixwidth}
\centering
\includegraphics[width=\textwidth]{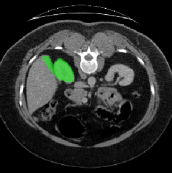}
\end{minipage}%
}%
\subfigure{
\begin{minipage}[t]{\patchsixwidth}
\centering
\includegraphics[width=\textwidth]{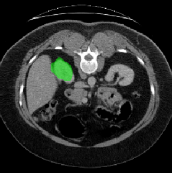}
\end{minipage}%
}%
\subfigure{
\begin{minipage}[t]{\patchsixwidth}
\centering
\includegraphics[width=\textwidth]{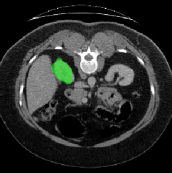}
\end{minipage}%
}%
\subfigure{
\begin{minipage}[t]{\patchsixwidth}
\centering
\includegraphics[width=\textwidth]{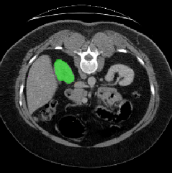}
\end{minipage}%
}%

\subfigure{
\begin{minipage}[t]{\patchsixwidth}
\centering
\includegraphics[width=\textwidth]{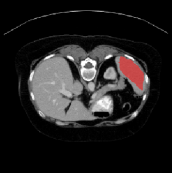}
\end{minipage}%
}%
\subfigure{
\begin{minipage}[t]{\patchsixwidth}
\centering
\includegraphics[width=\textwidth]{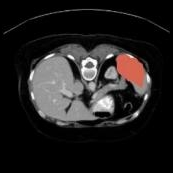}
\end{minipage}%
}%
\subfigure{
\begin{minipage}[t]{\patchsixwidth}
\centering
\includegraphics[width=\textwidth]{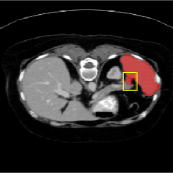}
\end{minipage}%
}%
\subfigure{
\begin{minipage}[t]{\patchsixwidth}
\centering
\includegraphics[width=\textwidth]{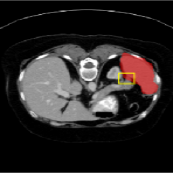}
\end{minipage}%
}%
\subfigure{
\begin{minipage}[t]{\patchsixwidth}
\centering
\includegraphics[width=\textwidth]{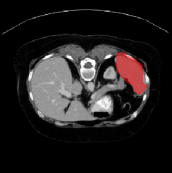}
\end{minipage}%
}%
\subfigure{
\begin{minipage}[t]{\patchsixwidth}
\centering
\includegraphics[width=\textwidth]{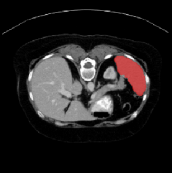}
\end{minipage}%
}%

\subfigure{
\begin{minipage}[t]{\patchsixwidth}
\centering
\includegraphics[width=\textwidth]{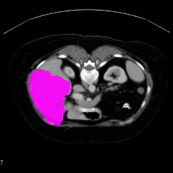}
\end{minipage}%
}%
\subfigure{
\begin{minipage}[t]{\patchsixwidth}
\centering
\includegraphics[width=\textwidth]{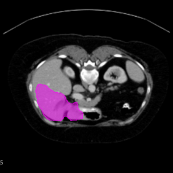}
\end{minipage}%
}%
\subfigure{
\begin{minipage}[t]{\patchsixwidth}
\centering
\includegraphics[width=\textwidth]{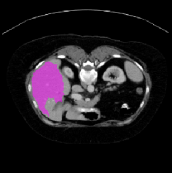}
\end{minipage}%
}%
\subfigure{
\begin{minipage}[t]{\patchsixwidth}
\centering
\includegraphics[width=\textwidth]{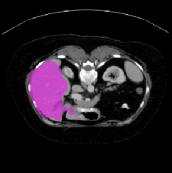}
\end{minipage}%
}%
\subfigure{
\begin{minipage}[t]{\patchsixwidth}
\centering
\includegraphics[width=\textwidth]{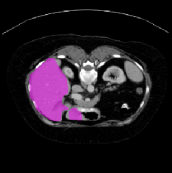}
\end{minipage}%
}%
\subfigure{
\begin{minipage}[t]{\patchsixwidth}
\centering
\includegraphics[width=\textwidth]{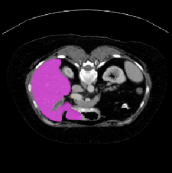}
\end{minipage}%
}%
\captionsetup{justification=raggedright}
\caption{Segmentation results using Setting1 on the SABS dataset. From left to right: ALPNet, ADNet, Q-Net, CRAPNet, SQPFNet (Ours), and Ground Truth (GT). From top to bottom: Left Kidney (LK), Right Kidney (RK), Spleen, and Liver.}
\label{fig2}
\end{figure*}

\begin{figure*}[!t]
\setlength{\belowcaptionskip}{1mm}
\centering
\subfigure{
\begin{minipage}[t]{\patchsixwidth}
\centering
{\scriptsize{ALPNet\cite{b20}}}
\end{minipage}
\begin{minipage}[t]{\patchsixwidth}
\centering
{\scriptsize{ADNet\cite{b21}}}
\end{minipage}
\begin{minipage}[t]{\patchsixwidth}
\centering
{\scriptsize{Q-Net\cite{b4}}}
\end{minipage}
\begin{minipage}[t]{\patchsixwidth}
\centering
{\scriptsize{CRAPNet\cite{b2}}}
\end{minipage}
\begin{minipage}[t]{\patchsixwidth}
\centering
{\scriptsize{SQPFNet(Ours)}}
\end{minipage}
\begin{minipage}[t]{\patchsixwidth}
\centering
{\scriptsize{GT}}
\end{minipage}
}

\subfigure{
\begin{minipage}[t]{\patchsixwidth}
\centering
\includegraphics[width=\textwidth]{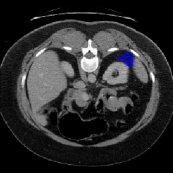}
\end{minipage}%
}%
\subfigure{
\begin{minipage}[t]{\patchsixwidth}
\centering
\includegraphics[width=\textwidth]{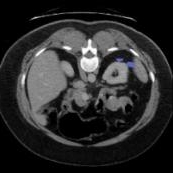}
\end{minipage}%
}%
\subfigure{
\begin{minipage}[t]{\patchsixwidth}
\centering
\includegraphics[width=\textwidth]{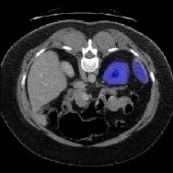}
\end{minipage}%
}%
\subfigure{
\begin{minipage}[t]{\patchsixwidth}
\centering
\includegraphics[width=\textwidth]{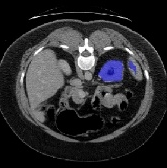}
\end{minipage}%
}%
\subfigure{
\begin{minipage}[t]{\patchsixwidth}
\centering
\includegraphics[width=\textwidth]{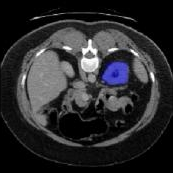}
\end{minipage}%
}%
\subfigure{
\begin{minipage}[t]{\patchsixwidth}
\centering
\includegraphics[width=\textwidth]{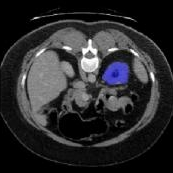}
\end{minipage}%
}%

\subfigure{
\begin{minipage}[t]{\patchsixwidth}
\centering
\includegraphics[width=\textwidth]{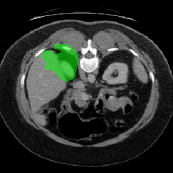}
\end{minipage}%
}%
\subfigure{
\begin{minipage}[t]{\patchsixwidth}
\centering
\includegraphics[width=\textwidth]{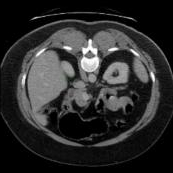}
\end{minipage}%
}%
\subfigure{
\begin{minipage}[t]{\patchsixwidth}
\centering
\includegraphics[width=\textwidth]{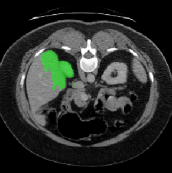}
\end{minipage}%
}%
\subfigure{
\begin{minipage}[t]{\patchsixwidth}
\centering
\includegraphics[width=\textwidth]{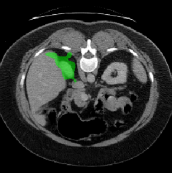}
\end{minipage}%
}%
\subfigure{
\begin{minipage}[t]{\patchsixwidth}
\centering
\includegraphics[width=\textwidth]{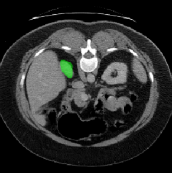}
\end{minipage}%
}%
\subfigure{
\begin{minipage}[t]{\patchsixwidth}
\centering
\includegraphics[width=\textwidth]{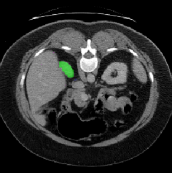}
\end{minipage}%
}%

\subfigure{
\begin{minipage}[t]{\patchsixwidth}
\centering
\includegraphics[width=\textwidth]{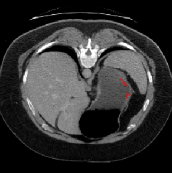}
\end{minipage}%
}%
\subfigure{
\begin{minipage}[t]{\patchsixwidth}
\centering
\includegraphics[width=\textwidth]{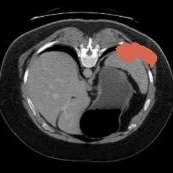}
\end{minipage}%
}%
\subfigure{
\begin{minipage}[t]{\patchsixwidth}
\centering
\includegraphics[width=\textwidth]{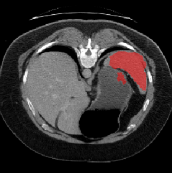}
\end{minipage}%
}%
\subfigure{
\begin{minipage}[t]{\patchsixwidth}
\centering
\includegraphics[width=\textwidth]{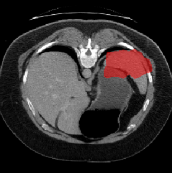}
\end{minipage}%
}%
\subfigure{
\begin{minipage}[t]{\patchsixwidth}
\centering
\includegraphics[width=\textwidth]{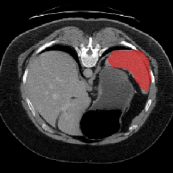}
\end{minipage}%
}%
\subfigure{
\begin{minipage}[t]{\patchsixwidth}
\centering
\includegraphics[width=\textwidth]{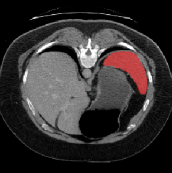}
\end{minipage}%
}%

\subfigure{
\begin{minipage}[t]{\patchsixwidth}
\centering
\includegraphics[width=\textwidth]{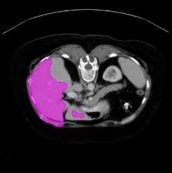}
\end{minipage}%
}%
\subfigure{
\begin{minipage}[t]{\patchsixwidth}
\centering
\includegraphics[width=\textwidth]{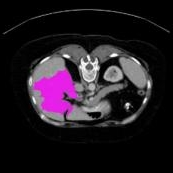}
\end{minipage}%
}%
\subfigure{
\begin{minipage}[t]{\patchsixwidth}
\centering
\includegraphics[width=\textwidth]{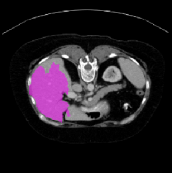}
\end{minipage}%
}%
\subfigure{
\begin{minipage}[t]{\patchsixwidth}
\centering
\includegraphics[width=\textwidth]{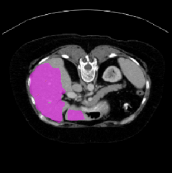}
\end{minipage}%
}%
\subfigure{
\begin{minipage}[t]{\patchsixwidth}
\centering
\includegraphics[width=\textwidth]{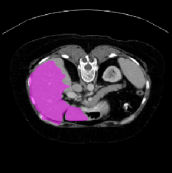}
\end{minipage}%
}%
\subfigure{
\begin{minipage}[t]{\patchsixwidth}
\centering
\includegraphics[width=\textwidth]{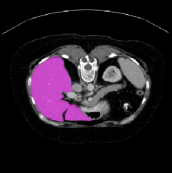}
\end{minipage}%
}%
\captionsetup{justification=raggedright}
\caption{Segmentation results using Setting2 on the SABS dataset. From left to right: ALPNet, ADNet, Q-Net, CRAPNet, SQPFNet (Ours), and GT(Ground Truth). From top to bottom: Left Kidney (LK), Right Kidney (RK), Spleen, and Liver.}
\label{fig3}
\end{figure*}

\begin{figure*}[t]
\setlength{\belowcaptionskip}{5mm}
\centering
\subfigure{
\begin{minipage}[t]{\patchfivewidth}
\centering
{\scriptsize{ADNet\cite{b21}}}
\end{minipage}
\begin{minipage}[t]{\patchfivewidth}
\centering
{\scriptsize{Q-Net\cite{b4}}}
\end{minipage}
\begin{minipage}[t]{\patchfivewidth}
\centering
{\scriptsize{CRAPNet\cite{b2}}}
\end{minipage}
\begin{minipage}[t]{\patchfivewidth}
\centering
{\scriptsize{SQPFNet(Ours)}}
\end{minipage}
\begin{minipage}[t]{\patchfivewidth}
\centering
{\scriptsize{GT}}
\end{minipage}
}

\subfigure{
\begin{minipage}[t]{\patchfivewidth}
\centering
\includegraphics[width=\textwidth]{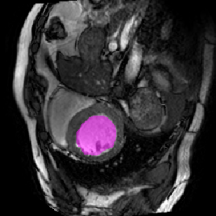}
\end{minipage}%
}%
\subfigure{
\begin{minipage}[t]{\patchfivewidth}
\centering
\includegraphics[width=\textwidth]{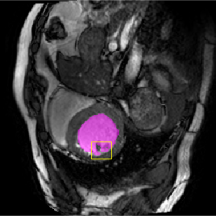}
\end{minipage}%
}%
\subfigure{
\begin{minipage}[t]{\patchfivewidth}
\centering
\includegraphics[width=\textwidth]{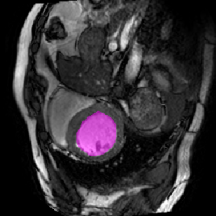}
\end{minipage}%
}%
\subfigure{
\begin{minipage}[t]{\patchfivewidth}
\centering
\includegraphics[width=\textwidth]{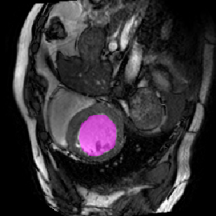}
\end{minipage}
}%
\subfigure{
\begin{minipage}[t]{\patchfivewidth}
\centering
\includegraphics[width=\textwidth]{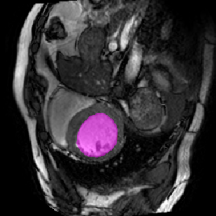}
\end{minipage}%
}%

\subfigure{
\begin{minipage}[t]{\patchfivewidth}
\centering
\includegraphics[width=\textwidth]{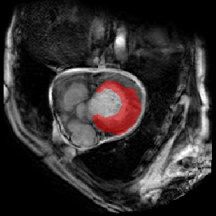}
\end{minipage}%
}%
\subfigure{
\begin{minipage}[t]{\patchfivewidth}
\centering
\includegraphics[width=\textwidth]{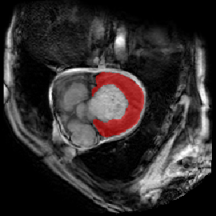}
\end{minipage}%
}%
\subfigure{
\begin{minipage}[t]{\patchfivewidth}
\centering
\includegraphics[width=\textwidth]{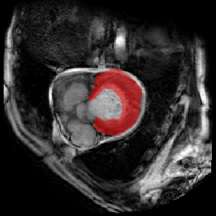}
\end{minipage}%
}%
\subfigure{
\begin{minipage}[t]{\patchfivewidth}
\centering
\includegraphics[width=\textwidth]{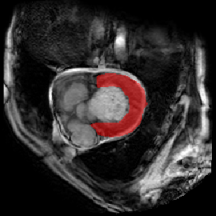}
\end{minipage}%
}%
\subfigure{
\begin{minipage}[t]{\patchfivewidth}
\centering
\includegraphics[width=\textwidth]{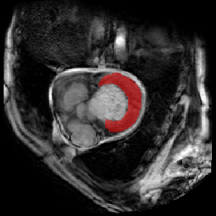}
\end{minipage}%
}%

\subfigure{
\begin{minipage}[t]{\patchfivewidth}
\centering
\includegraphics[width=\textwidth]{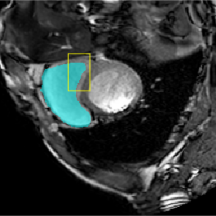}
\end{minipage}%
}%
\subfigure{
\begin{minipage}[t]{\patchfivewidth}
\centering
\includegraphics[width=\textwidth]{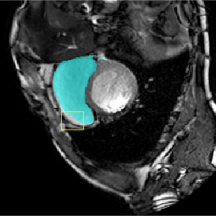}
\end{minipage}%
}%
\subfigure{
\begin{minipage}[t]{\patchfivewidth}
\centering
\includegraphics[width=\textwidth]{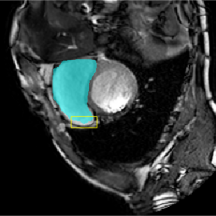}
\end{minipage}%
}%
\subfigure{
\begin{minipage}[t]{\patchfivewidth}
\centering
\includegraphics[width=\textwidth]{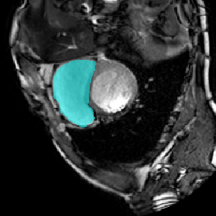}
\end{minipage}%
}%
\subfigure{
\begin{minipage}[t]{\patchfivewidth}
\centering
\includegraphics[width=\textwidth]{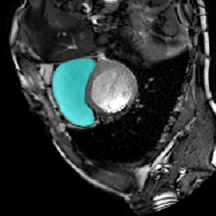}
\end{minipage}%
}%
\captionsetup{justification=raggedright}
\caption{Comparison of segmentation results from Setting1 on CMR dataset. From left to right: ADNet, Q-Net, CRAPNet, SQPFNet(Ours) and GT(Ground Truth). From top to bottom: LY-BP, LV-MYO and RV.}
\label{fig4}
\end{figure*}

We conducted comparisons of our methods with ALPNet\cite{b20}, ADNet\cite{b21}, QNet\cite{b4}, AAS-DCL\cite{b27}, and CRAPNet\cite{b2} on both the SABS\cite{b25} and CMR\cite{b26} datasets. To ensure a fair comparison, we evaluated them under two settings: Setting 1, where the test class may appear during the training phase, and Setting 2, where the test class is not visible to the model at all, following the approach of SSL-APLNet. For the CMR datasets, since organs consistently appear together, it's not feasible to exclude the test class from the model's training stage. Therefore, we only consider Setting 1 for the CMR datasets. 

Table \ref{table1} presents the model and other methods' performance on the SABS dataset, while Table \ref{table2} displays their performance on the CMR dataset. Our model exhibits superior performance across both settings, as illustrated in Table \ref{table1}. Specifically, under Setting 1, our model achieves an average Dice score of 77.00\%. Notably, in liver organ segmentation, our model outperforms ALPNet and QNet by 6.50\% and 4.09\%, respectively. 

Figure \ref{fig2} illustrates the specific segmentation results on the SABS dataset under Setting 1. Our model demonstrates higher accuracy in segmentation and effectively reduces unnecessary segmentation. Even under Setting 2, where the test class is entirely invisible to the model, our model performs well, as depicted in Figure \ref{fig3}. Given that SABS is a multi-organ dataset with many organs of small size, precise segmentation proves challenging. However, our model excels, particularly in the scenario of completely invisible classes, which closely resembles real-world conditions. Figure \ref{fig3} highlights our model's ability to accurately segment very small target organs, such as the right kidney. In contrast, the ADNet method fails to produce effective segmentation predictions, and other methods exhibit varying degrees of over-segmentation. On the CMR dataset, as shown in Table \ref{table2}, our model generally outperforms others. 

Figure \ref{fig4} provides a visual representation of our model's segmentation, indicating its superior accuracy, particularly in the right ventricle (RV) organ. While other methods display complete segmentation, our segmentation aligns more closely with the real mask.

\subsection{Ablation Study}
In order to verify the effectiveness of our proposed method, we conducted ablation experiments on different modules, and the relevant experimental results are shown in Table \ref{table3}. All ablation experiments were conducted on the SABS dataset at setting 2, and we also listed the results of 5 cross-validation sets in Table \ref{table3}. 

We first consider the impact of supporting prototypes on model performance. The traditional way is to build a class prototype for each class. The way we use is to divide the foreground area, build a class prototype for each part, and then average the prototypes of different parts to get the final class prototype. We call our approach multiple supporting prototypes. As you can see from the table, building multiple support prototypes is a significant improvement over building a single support prototype, with an increase of 2.68\%. In addition, we built a query prototype following the way we built a support prototype. As you can see from the table, building query prototypes on top of multiple supported prototypes improved the performance of the model by 0.9\%. Finally, through the combination of multiple support prototypes and query prototypes, the performance of the model has been greatly improved compared with the single support prototype, from 66.41\% to 69.87\%. 

\begin{figure*}[!t]
\setlength{\belowcaptionskip}{5mm}
\centering
\subfigure{
\begin{minipage}[t]{\patchsevenwidth}
\centering
{\scriptsize{SupportImg}}
\end{minipage}
\begin{minipage}[t]{\patchsevenwidth}
\centering
{\scriptsize{QueryImg}}
\end{minipage}
\begin{minipage}[t]{\patchsevenwidth}
\centering
{\scriptsize{SSP}}
\end{minipage}
\begin{minipage}[t]{\patchsevenwidth}
\centering
{\scriptsize{MSP}}
\end{minipage}
\begin{minipage}[t]{\patchsevenwidth}
\centering
{\scriptsize{SSP+QP}}
\end{minipage}
\begin{minipage}[t]{\patchsevenwidth}
\centering
{\scriptsize{MSP+QP}}
\end{minipage}
\begin{minipage}[t]{\patchsevenwidth}
\centering
{\scriptsize{GT}}
\end{minipage}
}

\subfigure{
\begin{minipage}[t]{\patchsevenwidth}
\centering
\includegraphics[width=\textwidth]{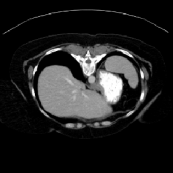}
\end{minipage}%
}%
\subfigure{
\begin{minipage}[t]{\patchsevenwidth}
\centering
\includegraphics[width=\textwidth]{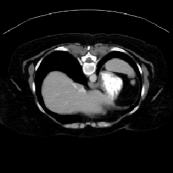}
\end{minipage}%
}%
\subfigure{
\begin{minipage}[t]{\patchsevenwidth}
\centering
\includegraphics[width=\textwidth]{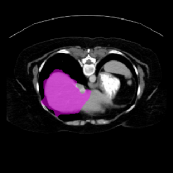}
\end{minipage}%
}%
\subfigure{
\begin{minipage}[t]{\patchsevenwidth}
\centering
\includegraphics[width=\textwidth]{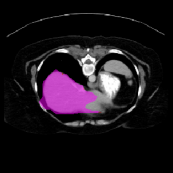}
\end{minipage}%
}%
\subfigure{
\begin{minipage}[t]{\patchsevenwidth}
\centering
\includegraphics[width=\textwidth]{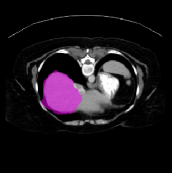}
\end{minipage}%
}%
\subfigure{
\begin{minipage}[t]{\patchsevenwidth}
\centering
\includegraphics[width=\textwidth]{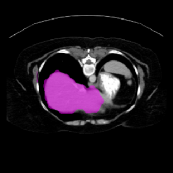}
\end{minipage}%
}%
\subfigure{
\begin{minipage}[t]{\patchsevenwidth}
\centering
\includegraphics[width=\textwidth]{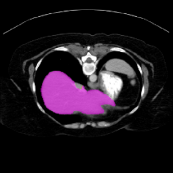}
\end{minipage}%
}%

\subfigure{
\begin{minipage}[t]{\patchsevenwidth}
\centering
\includegraphics[width=\textwidth]{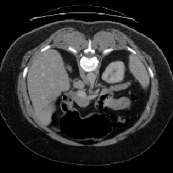}
\end{minipage}%
}%
\subfigure{
\begin{minipage}[t]{\patchsevenwidth}
\centering
\includegraphics[width=\textwidth]{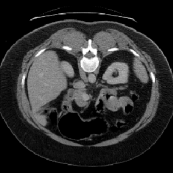}
\end{minipage}%
}%
\subfigure{
\begin{minipage}[t]{\patchsevenwidth}
\centering
\includegraphics[width=\textwidth]{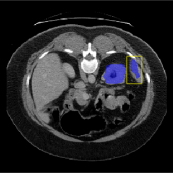}
\end{minipage}%
}%
\subfigure{
\begin{minipage}[t]{\patchsevenwidth}
\centering
\includegraphics[width=\textwidth]{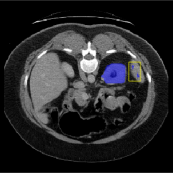}
\end{minipage}%
}%
\subfigure{
\begin{minipage}[t]{\patchsevenwidth}
\centering
\includegraphics[width=\textwidth]{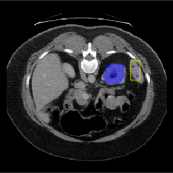}
\end{minipage}%
}%
\subfigure{
\begin{minipage}[t]{\patchsevenwidth}
\centering
\includegraphics[width=\textwidth]{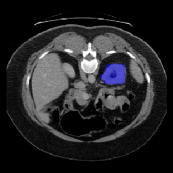}
\end{minipage}%
}%
\subfigure{
\begin{minipage}[t]{\patchsevenwidth}
\centering
\includegraphics[width=\textwidth]{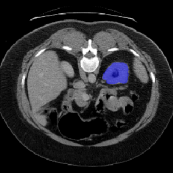}
\end{minipage}%
}%

\captionsetup{justification=raggedright}
\caption{Predictions generated by various prototypes. From left to right: support image, query image, predictions of SSP, MSP, SSP+QP, MSP+QP, and GT (Ground Truth), where SSP denotes single support prototype, MSP denotes multiple support prototypes, QP denotes query prototype, and GT denotes Ground Truth.}
\label{fig5}
\end{figure*}

The predictions produced by the different prototypes are shown in Figure \ref{fig5}. Among them, in the segmentation of liver organs, it can be seen that the use of a single support prototype cannot produce a comprehensive segmentation, and the use of multiple support prototypes has significantly improved the effect, and the combination of multiple support prototypes and query prototypes can produce a more comprehensive segmentation. On the left kidney, the use of a single supporting prototype will produce a relatively large amount of over-segmentation, which will be reduced after the introduction of the query prototype, but some pixels are still over-segmented. However, the combination of multiple support prototypes and query prototypes can avoid over-segmentation.

\begin{table*}[t]
\centering
\captionsetup{justification=raggedright}
\caption{Ablation study results of different modules on the SABS dataset under setting 2.}
\resizebox{\textwidth}{!}{
\begin{tabular}{@{}ccc|ccccc|c@{}}
\toprule
\textbf{SSP} & \textbf{MSP} & \textbf{QP} & \textbf{Flod1} & \textbf{Fold2} & \textbf{Fold3} & \textbf{Fold4} & \textbf{Fold5} & \textbf{Mean Dice Score$\uparrow$} \\ 
\midrule
\checkmark   &              &       
& \color{blue}{68.95}          & 55.24    & 65.79   & \textbf{\color{red}{75.39}} & 66.67       & 66.41\\
             & \checkmark   &       
& \textbf{\color{red}{69.65}} & \color{blue}{59.95}    & \color{blue}{68.76}   & \color{blue}{73.57}          & \color{blue}{73.54}       & \color{blue}{69.09} \\
\checkmark   &              & \checkmark
& 68.63      & 56.74        & 67.69   & 73.56          & 65.83       & 66.49 \\
             & \checkmark   & \checkmark
& 68.57      & \textbf{\color{red}{61.94}} & \textbf{\color{red}{72.04}}         & 72.99       & \textbf{\color{red}{73.80}} & \textbf{\color{red}{69.87}} \\ 
\bottomrule
\end{tabular}
}\label{table3}
\end{table*}

\begin{figure*}[t]
\setlength{\belowcaptionskip}{5mm}
\centerline{\includegraphics[width=\textwidth]{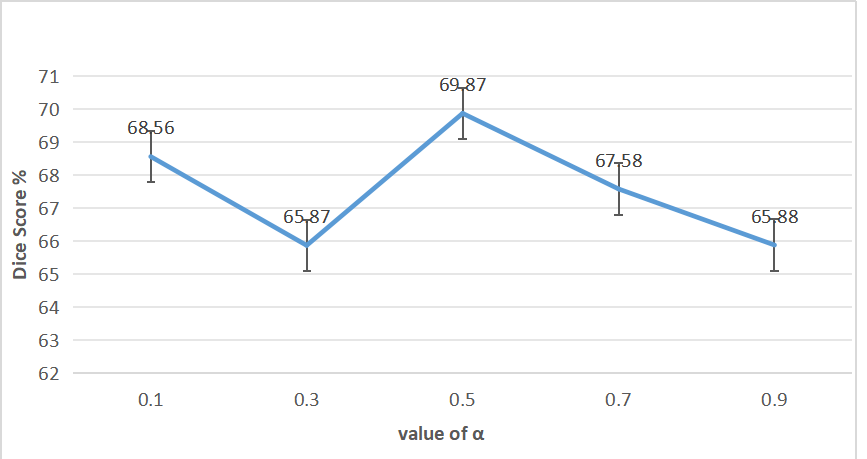}}
\captionsetup{justification=raggedright}
\caption{Ablation study results of ${\alpha}$.}
\label{fig6}
\end{figure*}

Furthermore, we conducted an ablation experiment on the combination coefficient ${\alpha}$, which represents the balance between the support prototype and the query prototype. The results are depicted in Figure \ref{fig6}. It's evident from the figure that as ${\alpha}$ decreases, indicating a gradual introduction of the query prototype, the model's performance improves. The optimal performance of 69.87\% is achieved when ${\alpha}$ is set to 0.5. However, further decreasing ${\alpha}$ beyond this point leads to a decline in performance, reaching a sub-optimal level of 68.56\% when ${\alpha}$ is reduced to 0.1. 

Moreover, we compared the number of model parameters and floating-point calculation times across different methods, with a unified input size of 256×256. The experimental results are summarized in Table~\ref{table4}. Given the similarity in the model backbone across methods, the number of parameters is comparable. Notably, our model stands out as a relatively lightweight model. Despite the introduction of the query prototype, our model achieves improved performance without increasing the number of model parameters.

\begin{table}[]
\centering
\captionsetup{justification=raggedright}
\caption{Model size and computational complexity of various models.}
\begin{tabularx}{0.85\textwidth}{m{0.3\textwidth} m{0.2\textwidth} m{0.35\textwidth} }
\toprule
\textbf{Methods} & \textbf{Parameters$\downarrow$} & ${\approx}$ \textbf{GFLOPs$\downarrow$} \\ \midrule
ALPNet\cite{b20}            & 43.02~M             & 90.17442      \\
ADNet\cite{b21}             & 43.02~M             & 90.17439      \\
QNet\cite{b4}              & 46.12~M              & 266.86460     \\
CRAPNet\cite{b2}           & 44.01~M              & 91.51657      \\
SQPFNet(Ours)            & 43.02~M             & 90.17439      \\ 
\bottomrule
\end{tabularx}
\label{table4}
\end{table}

\section{Conclusion}
In this paper, we propose a new Support-Query Prototype Fusion network (SQPFNet) to alleviate the problems of intra-class and inter-class gaps in few-shot segmentation. Our model first divides the foreground area of the support picture into multiple parts, constructs prototypes for each part, and then averages the multiple part prototypes to get the support prototypes and the coarse mask predictions generated by the support prototypes. Then, the query prototype is constructed according to the pattern of the prototypical network by using the query features and the segmented coarse mask. Ultimately, we implement weighted combination support and query prototypes to build high quality prototypes and complete the segmentation task.

\backmatter

\section{Declarations}
\subsection{Conflict of interest}
The authors declare that they have no known competing financial interests or personal relationships that could have appeared to influence the work reported in this paper.
\subsection{Ethical Approval}
Not applicable.
\subsection{Funding}
Natural Science Foundation of China under Grant 62372190 and Industry University Cooperation Project of Fujian Province under Grant 2021H6030.
\subsection{Availability of data and materials}
The datasets we use are all public open-source data,
they can be obtained from \cite{b25,b26}









\end{document}